\documentclass[journal,twoside,web]{ieeecolor}
\usepackage{generic}
\usepackage{cite}
\usepackage{amsmath,amssymb,amsfonts}
\usepackage{algorithm}
\usepackage{algpseudocode}
\usepackage{graphicx}
\usepackage{subcaption}
\usepackage{textcomp}
\usepackage{multirow, booktabs}

\def\BibTeX{{\rm B\kern-.05em{\sc i\kern-.025em b}\kern-.08em
    T\kern-.1667em\lower.7ex\hbox{E}\kern-.125emX}}
\begin{document}
\title{Text-to-CAD Retrieval: a Strong Baseline}
\author{Honghu Pan, Zibo Du, Daxiang Liu, Chengliang Liu, Xiaoling Luo
\thanks{H. Pan is with School of Artificial Intelligence and Robotics, Hunan University. (Email: honghupan@hnu.edu.cn)}
\thanks{Z. Du, D. Liu, and X. Luo are with  College of Computer Science and Software Engineer, Shenzhen University. (Email: 2410103099@mails.szu.edu.cn,  2500101036@mails.szu.edu.cn, and xlluo@szu.edu.cn)}
\thanks{C. Liu is with  Faculty of Science and Technology, University of Macau. (Email: liucl1996@163.com)}
}

\maketitle

\begin{abstract}
Text-based retrieval of Computer-Aided Design (CAD) models is a critical yet underexplored task for the reuse of legacy industrial designs.
Existing CAD repositories are typically searched using filenames or directories, which limits the efficiency, scalability, and accuracy of design retrieval.
In this paper, we formally introduce text-to-CAD retrieval as a new cross-modal retrieval task, aiming to retrieve semantically relevant CAD models from large-scale databases given natural language queries.
Leveraging paired text-CAD annotations from the Text2CAD dataset, we establish a practical benchmark for this task.
To achieve text-based retrieval, we propose a unified framework that learns multi-modal CAD embeddings from both procedural sequences and geometric point clouds.
Specifically, a sequence encoder captures the construction logic of CAD models, while a point encoder extracts explicit geometric features.
A text encoder is used to learn semantic representations of textual queries.
During training, we introduce a novel feature decoder that reconstructs masked sequence features via cross-attention with text and point features, encouraging implicit multi-modal alignment.
At inference time, we remove this auxiliary decoder to enable efficient retrieval using concatenated sequence-point features.
Our framework serves as a strong baseline for text-to-CAD retrieval and lays the foundation for downstream CAD generation paradigms, such as retrieval-augmented generation.
The source code will be released.
\end{abstract}


\section{Introduction}
\label{sec:introduction}
\begin{figure*}
	\includegraphics[width=0.99 \textwidth]{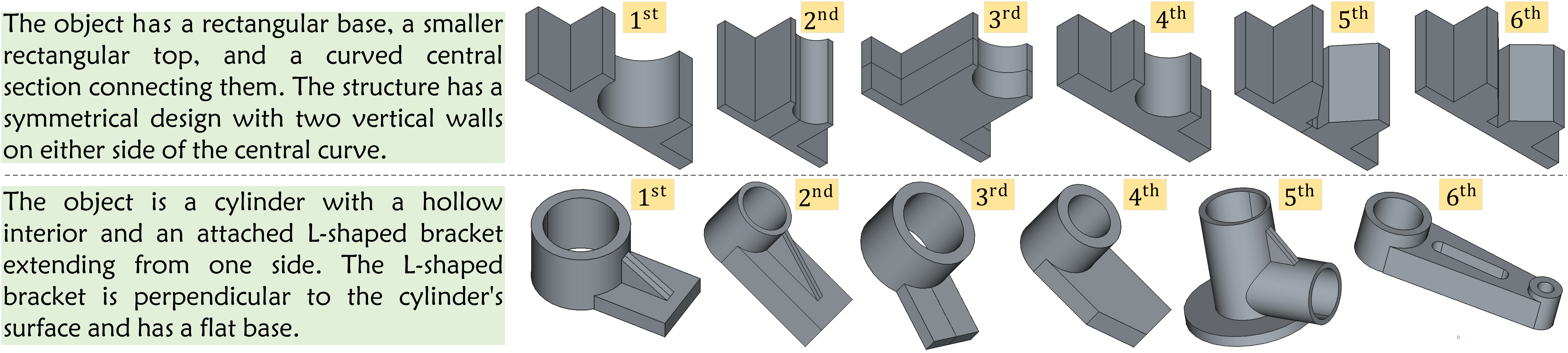}
	\vspace{-0.15cm} 
	\caption{
		Visualization of retrieved CAD models for given text queries. The proposed text-to-CAD model retrieves samples semantically associated with each query, facilitating industrial legacy reuse and reducing human effort.
	}
	\label{fig:retrieval}
\end{figure*}

\IEEEPARstart{A}{s} a fundamental tool in modern engineering and industry, Computer-Aided Design (CAD) enables the creation of precise, manufacturable 3D models for engineering analysis and production.
Despite its critical role, traditional CAD part design remains highly labor-intensive, relying heavily on repetitive and manual operations performed by experienced mechanical engineers.
Meanwhile, large enterprises have accumulated vast repositories of historical CAD models over years of development.
Reusing these designs typically depends on filenames, directory structures, or personal memory, leading to inefficient and error-prone retrieval that does not scale well with database size.
Despite these practical needs, there is currently no dedicated retrieval paradigm that enables engineers to query large-scale CAD repositories using natural language descriptions.
As illustrated in Fig.~\ref{fig:retrieval}, we therefore formalize \emph{text-to-CAD retrieval} as a new task, which aims to retrieve semantically relevant CAD models from large databases given textual queries.

To enable cross-modal matching~\cite{ACD,t2i,tii2}, text-to-CAD retrieval seeks to project textual descriptions and CAD models into a shared embedding space, where matched pairs are close and mismatched pairs are far apart.
A key challenge lies in representing CAD models in a form amenable to learning and retrieval.
Inspired by recent advances in CAD generation~\cite{Complexgen,Brepgen,CMT,Deepcad,SkexGen,HNC-CAD,Diffusion-CAD,tii1,tii3}, CAD models are commonly represented as sequences of discrete commands and parameters, mimicking how designers create models step-by-step.
For example, Text2CAD~\cite{Text2cad}, a pioneering work in text-to-CAD generation, adopts a sketch-and-extrude paradigm and represents CAD models as sequences of 2D coordinates. The resulting text-driven CAD generation datasets~\cite{Text2cad,Cadtranslator}, which contain paired text-sequence annotations, serve as a natural testbed for exploring text-to-CAD retrieval.
However, while such sequence representations contain rich procedural information, they may overlook explicit geometric characteristics that are critical for text-to-CAD retrieval.

In this work, we propose a unified text-to-CAD retrieval framework that leverages multi-modal representations of CAD models, including procedural sequences provided by CAD generation datasets and geometric point clouds extracted from CAD entities.
To learn discriminative multi-modal representations, we implement three separate encoders: a text encoder comprising a pretrained language tokenizer~\cite{CLIP,BERT} followed by a trainable Transformer encoder; a sequence encoder built with a stack of Transformer layers; and a point encoder adopted from existing point-based networks~\cite{PointNet,PointTransformer1,PointTransformer2}.
In the training stage, we compute the InfoNCE loss separately for two modality pairs: text–sequence and text–point. Notably, the sequence branch and the point cloud branch are trained independently without any cross-branch interaction.
In the inference stage, the final CAD representation is formed by concatenating sequence and point features, while the text representation is duplicated to match the CAD embedding dimension, enabling similarity computation in a shared space.

During training, we introduce a novel feature decoder to encourage \textbf{implicit multi-modal alignment} by reconstructing sequence features from text and point features.
Specifically, this decoder is composed of multiple cross-attention layers, where masked sequence features with stop gradient serve as the query, while text features and point cloud features alternately act as the key and value.
We then compute a mean square error (MSE) loss between the reconstructed sequence features and their ground-truth counterparts. This loss drives gradients to the text and point encoders to enforce cross-modal alignment.
This auxiliary module is used solely for model training and is discarded at inference time, enabling fast and efficient cross-modal retrieval.

We test the proposed approach on two text-to-CAD generation datasets: Text2CAD~\cite{Text2cad} and CadTranslator~\cite{Cadtranslator}.
They share the same CAD source (DeepCAD~\cite{Deepcad}), and obtain language descriptions by Large Language Models (LLMs).
Experimental results strongly support the effectiveness of our multi-modal design. For example, on Text2CAD, the Rank1 accuracy improves from 5.00 when using only sequence features to 9.71 after incorporating multi-modal representations and alignment.
The contributions of this paper are summarized as follows.
First, leveraging the paired text-CAD annotations from text-to-CAD generation datasets~\cite{Text2cad,Cadtranslator}, we formally introduce \emph{text-to-CAD retrieval} as a new cross-modal retrieval task.
This task is of practical importance for CAD model reuse in industrial design workflows, and also serves as a foundational technique for downstream CAD generation paradigms, such as retrieval-augmented generation (RAG)~\cite{RAG1,RAG2,RAG3}.
Second, we establish a strong baseline for text-to-CAD retrieval by learning multi-modal CAD representations from procedural sequences and geometric point clouds, which incorporates a novel feature decoder to achieve implicit multi-modal alignment during training.
Third, quantitative analyses demonstrate the superiority of the proposed approach over existing text-to-3D shape retrieval models, while qualitative results show that our method exhibits sensitivity to geometric descriptions in the text.

The remainder of this paper is organized as follows.
Section~\ref{sec:relate} summarizes recent studies related to this paper;
Section~\ref{sec:method} elaborates the framework of the proposed approach;
Section~\ref{sec:experi} presents the qualitative results and quantitative analyses;
finally, Section~\ref{sec:conclu} draws conclusions and discussions.

\section{Related Works}
\label{sec:relate}

\subsection{Deep Learning for CAD Models}
As a cornerstone of modern engineering and manufacturing, CAD has drawn considerable attention from the machine learning and computer vision communities due to its practical industrial significance.
Recent research has mainly focused on generative models for CAD, which typically leverage three paradigms to represent CAD models: boundary representation (B-rep)~\cite{Complexgen,Brepgen,CMT}, sequence representation~\cite{Deepcad,SkexGen,HNC-CAD,Diffusion-CAD,RECAD}, and text representation~\cite{CAD-Llama,FlexCAD} for fine-tuning LLMs~\cite{LLaMA3,Gpt-4}.
B-rep describes 3D CAD entities by explicitly defining object boundaries with a set of surfaces and curves, providing an accurate geometric shape description.
In contrast, sequence representation mimics human design procedures by representing CAD construction as a sequence of modeling commands and parameters.
Meanwhile, text representation transforms CAD sequences into textual descriptions or Python codes that align with the output format of LLMs.

This paper adopts a sequence-based representation for CAD modeling.
In this area, early studies primarily focused on unconditional generation, which aims to randomly generate samples that follow the same distribution as the given data.
DeepCAD~\cite{Deepcad} was the pioneering study that constructs CAD models as sequences of discrete commands and associated quantized parameters, employing an autoregressive model to learn their distributions.
SkexGen~\cite{SkexGen} and HNC-CAD~\cite{HNC-CAD} further transformed CAD models into tree-like structures for hierarchical modeling, utilizing VQ-VAE~\cite{VQ-VAE} to compress CAD models into discrete codebook indices. 
Diffusion-CAD~\cite{Diffusion-CAD} leveraged diffusion models to improve sample diversity and controllability.

Recent studies have proposed the text-guided CAD generation task, which aims to produce CAD models that semantically match a given language description. Text2CAD~\cite{Text2cad} and CadTranslator~\cite{Cadtranslator} each provided a text-to-CAD dataset, both of which use DeepCAD~\cite{Deepcad} as the base CAD repository.
For acquiring textual descriptions, CadTranslator first rendered multi-view CAD images using PythonOCC and then leveraged CoCa~\cite{CoCa} to describe the perspective images.
Text2CAD~\cite{Text2cad} first employed LLaVA-NeXT~\cite{Llavanext} to generate abstract shapes from multi-view images and then obtained multi-level descriptions using Mixtral-50B~\cite{Mixtral}.
Its descriptions are organized into four levels: abstract, beginner, intermediate, and expert, providing progressively more detailed CAD descriptions.

\subsection{Text-to-3D Retrieval}
Text-to-3D shape retrieval aims to retrieve 3D models from a database based on natural language queries.
Text2Shape~\cite{text2shape} first provided a benchmark by introducing the Text2Shape dataset, which contains natural language descriptions paired with ShapeNet models, and learned a shared latent space for text and voxelized 3D shapes.
Y²Seq2Seq~\cite{Y2Seq2Seq} proposed a view-based approach that renders multiple views of a 3D model and learns cross-modal representations through coupled sequence-to-sequence networks.
Parts2Words~\cite{parts2words} enabled fine-grained matching between words in text and parts in 3D shapes; however, it requires segmentation annotations.
To enhance retrieval performance, COM3D exploited cross-view correspondence and cross-modal mining using a scene representation transformer.
TriCoLo~\cite{Tricolo} introduced a trimodal contrastive learning framework that jointly aligns text, multi-view images, and 3D voxel representations. By incorporating image features as an intermediate modality, this approach reduces the modality gap between textual descriptions and 3D shapes.
To address the limited scale of paired text-shape data, SCA3D~\cite{sca3d} proposed augmenting training data by automatically generating captions for segmented 3D components using vision-language models.
To overcome the noise or disorder of point clouds and the ambiguity or incompleteness of texts, RoMa~\cite{roma} implemented two key modules: a Dual Attention Perception module and a Robust Negative Contrastive Learning module.

\subsection{Cross-Modal Retrieval using Multi-Modal Data}
Cross-modal retrieval encompasses a range of downstream tasks, including text-to-vision, text-to-motion, visible-to-infrared, and text-to-3D shape retrieval, among others.
A general paradigm for cross-modal retrieval is to learn a shared embedding space, where positive pairs are pulled closer and negative pairs are pushed apart.
However, unlike single-modal retrieval, cross-modal retrieval suffers from a significant modality gap, making it prone to misalignment between heterogeneous data sources.
Consequently, many approaches seek to leverage multi-modal data to bridge this gap.
The fundamental advantage of multi-modal data lies in its ability to provide multi-perspective and complementary information, thereby creating richer constraints for cross-modal alignment and enabling models to learn more robust and modality-agnostic shared representations.
For text-to-video retrieval, TEFAL~\cite{TEFAL} and EclipSE~\cite{Eclipse} incorporate audio as complementary information to reduce the discrepancy between videos and texts. For visible-to-infrared person re-identification, ACD~\cite{ACD} generates intermediate-modality images for both visible and infrared inputs, ensuring that each sample shares a common modal representation.
For text-to-motion retrieval, LAVIMO~\cite{LAVIMO}animates avatars from human motion and renders RGB videos to enable tri-modal retrieval.
For text-to-3D shape retrieval, both TriCoLo~\cite{Tricolo} and TriModal~\cite{TriModal} render multi-view images from 3D shapes as supplementary visual inputs.
In this paper, we propose to leverage both procedural sequences and geometric point clouds to match textual descriptions.

\begin{figure*}[ht]
	\includegraphics[width=0.99 \textwidth]{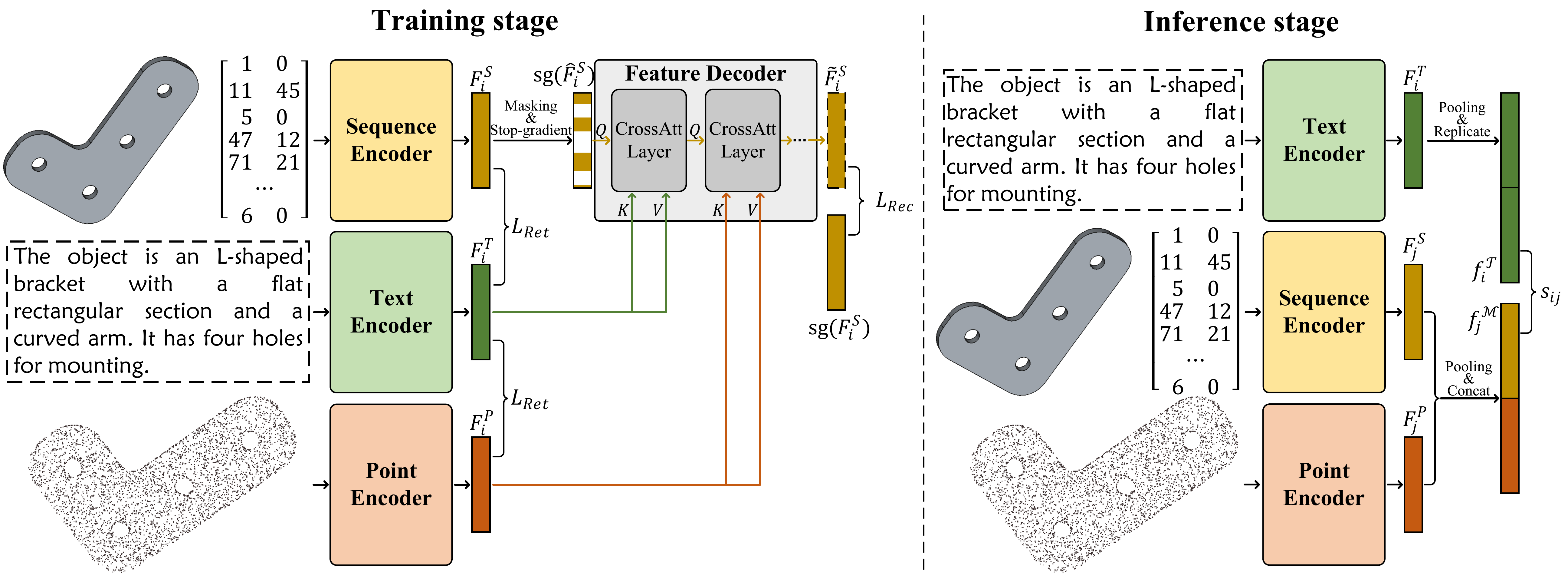}
	\vspace{-0.15cm} 
	\caption{
		Architecture of the proposed text-to-CAD retrieval pipeline.
		During training, modality-specific encoders extract sequence, text, and point cloud features, while a feature decoder reconstructs masked sequence features through cross-attention with text and point features.
		During inference, the decoder is discarded, and retrieval is performed by computing similarities between concatenated sequence–point features for CAD models and duplicated text features.
	}
	\label{fig:framework}
\end{figure*}

\section{Methods}
\label{sec:method}

\subsection{Problem Definition and Framework Overview}
\label{sec:pipeline}

We consider a dataset consisting of paired CAD models and textual descriptions.
Each CAD model is represented by a sequence of 2D geometric primitives, denoted as $S$, and an associated point cloud representation, denoted as $P$.
The corresponding textual description is denoted as $T$.
The objective of text-to-CAD retrieval is to learn a shared embedding space in which a CAD model $\mathcal{M}_i=\{S_i, P_i\}$ and its corresponding text description $\mathcal{T}_i=\{T_i\}$ are mapped close to each other, while non-matching pairs are pushed away.

To this end, we propose a cross-modal retrieval pipeline, as illustrated in Fig.~\ref{fig:framework}, which consists of a text encoder $\phi_T$, a sequence encoder $\phi_S$, and a point encoder $\phi_P$.
These encoders map the input modalities into a common feature space:
\begin{equation}
	F^T_i = \phi_T(T_i), \quad
	F^S_i = \phi_S(S_i), \quad
	F^P_i = \phi_P(P_i),
	\label{eq:encoding}
\end{equation}
where $F^T_i \in \mathbb{R}^{L_T \times D}$, $F^S_i \in \mathbb{R}^{L_S \times D}$, and $ F^P_i \in \mathbb{R}^{N_P \times D}$ denote the representations of text, CAD sequence, and point cloud, respectively.
Here, $L_T$, $L_S$, and $N_P$ denote the text length, sequence length, and number of points, respectively.
To facilitate cross-modal alignment, we further introduce a feature reconstruction decoder $\psi_S$, which aims to recover the sequence feature $F^S$ from its masked counterpart $\hat{F}^S_i$, conditioned on the text and point cloud features:
\begin{equation}
	\tilde{F}^S_i = \psi_S(\text{sg}(\hat{F}^S_i), F^T_i, F^P_i),
	\label{eq:decoding}
\end{equation}
where $\hat{F}^S_i$ is obtained by applying zero-masking to $F^S_i$, and $\text{sg}(\cdot)$ denotes the stop-gradient operator that blocks gradients from the reconstruction loss flowing back into the sequence encoder $\phi_S$.

During training, the encoders $\phi_T$, $\phi_S$, $\phi_P$, and the feature decoder $\psi_S$ are jointly optimized in an end-to-end manner.
At inference time, the decoder $\psi_S$ is removed, and only the encoders are retained for representation learning.
Specifically, we concatenate the sequence and point cloud features to construct the final CAD representation $f^{\mathcal{M}}_i \in \mathbb{R}^{2D}$, and replicate the text feature to form the final text representation $f^{\mathcal{T}}_i \in \mathbb{R}^{2D}$:
\begin{equation}
	f^{\mathcal{M}}_i = [f^S_i; f^P_i], \quad 
	f^{\mathcal{T}}_i = [f^T_i; f^T_i],
	\label{eq:concat}
\end{equation}
where $f^T_i, f^S_i, f^P_i \in \mathbb{R}^{D}$ are obtained by applying average pooling to $F^T_i$, $F^S_i$, and $F^P_i$, respectively.

The similarity between a text query $\mathcal{T}_i$ and a CAD model $\mathcal{M}_j$ is computed as
\begin{equation}
	s_{ij} = \rho\bigl(f^{\mathcal{T}}_i, f^{\mathcal{M}}_j\bigr),
	\label{eq:distance}
\end{equation}
where $\rho(\cdot,\cdot)$ denotes the cosine similarity.
CAD models are then ranked according to their similarity scores.

In Section~\ref{sec:model}, we detail the structure and network flow of the three encoders and the feature decoder.
In Section~\ref{sec:loss}, we present the loss functions for model training.

\subsection{Individual Module Design}
\label{sec:model}

As depicted in Fig.~\ref{fig:framework}, the proposed architecture includes three encoders responsible for multi-modal representation extraction, along with a feature decoder that enables implicit multi-modal alignment. These modules are respectively elaborated below.

\textbf{Text Encoder.}
The text encoder $\phi_T$ extracts high-level semantic representations from natural language descriptions.
It consists of two core components: a pretrained language model for generating word-level embeddings, and a Transformer encoder for learning discriminative text-level features.
Specifically, we first tokenize the input text with a pretrained language tokenizer and derive contextualized token embeddings from a large-scale pretrained language model, such as CLIP~\cite{CLIP} or BERT~\cite{BERT}. These token embeddings are then fed into stacked Transformer encoder layers~\cite{Transformer} to capture global semantic interactions across the entire text sequence.
The fundamental building block of the Transformer layers is the self-attention mechanism, formulated as:
\begin{equation}
	\begin{split}
		Q = F^T W^Q, \ K &= F^T W^K, \ V = F^T W^V, \\
		\text{SelfAttn}(Q, K, &V) = \text{softmax}\left(\frac{QK^T}{\sqrt{d_k}}\right)V,
	\end{split}
	\label{eq:attention}
\end{equation}
where $d_k$ is the dimension of the query vector.
The output features from the Transformer encoder serve as the final text-level representations, which are further used for subsequent retrieval and reconstruction tasks.

\textbf{Sequence Encoder.}
The sequence encoder $\phi_S$ processes CAD models represented as ordered sequences of parametric commands and arguments.
Following prior works~\cite{Text2cad} on CAD sequence modeling, we first embed each command and its associated parameters into continuous vector embeddings, then process these embeddings by a stack of Transformer encoder layers to learn sequence feature $F^S$.
Specifically, each token in the input CAD sequence is encoded as a 256-dimensional one-hot vector, yielding a one-hot representation $S_0 \in \mathbb{R}^{L_S \times 2 \times 256}$, where the second dimension corresponds to the $x$ and $y$ coordinates, denoted as $S_0^x$ and $S_0^y$ respectively. The initial CAD sequence embedding $F^S_0$ is then computed as:
\begin{equation}
	F^S_0 = S_0^x W^x + S_0^y W^y + P,
\end{equation}
where $P$ denotes the positional encoding. The subsequent Transformer encoding process follows Eq.~(\ref{eq:attention}) and is omitted here for brevity.

\textbf{Point Encoder.}
The point encoder $\phi_P$ is designed to capture explicit geometric information from 3D point clouds sampled from CAD voxels.
We employ point-based neural networks, such as PointNet~\cite{PointNet} or PointTransformer~\cite{PointTransformer1,PointTransformer2}, to extract discriminative geometric features $F^P$ that are invariant to point ordering.
Specifically, the encoder first maps raw point coordinates into a high-dimensional feature space, followed by hierarchical feature aggregation to model local geometric patterns and global shape information.
Followed by this point-based network, we implement two fully-connected layers with a ReLU activation function for linear mapping.

\begin{table*}[t]
	\footnotesize
	\centering
	\caption{Performance comparison on Text2CAD and CadTranslator. We implement our model with different text tokenizers (BERT vs. CLIP) and point encoders (PointNet vs. PointTransformer). The best results are highlighted in \textbf{bold}, and ``PointTrans'' denotes the PointTransformer encoder.}
	\vspace{-0.1cm} 
	\begin{tabular}{ll|ccccccc|ccccccc}
		\toprule
		\multirow{2}{*}{$\phi_T$} & \multirow{2}{*}{$\phi_P$} & \multicolumn{7}{c|}{Text2CAD}                      & \multicolumn{7}{c}{CadTranslator}                       \\ \cmidrule{3-16} 
		&                        & R1$\uparrow$    & R2$\uparrow$    & R5$\uparrow$    & R10$\uparrow$   & R20$\uparrow$   & MedR$\downarrow$  & Rsum$\uparrow$  & R1$\uparrow$    & R2$\uparrow$    & R5$\uparrow$    & R10$\uparrow$   & R20$\uparrow$   & MedR$\downarrow$  & Rsum$\uparrow$  \\ \midrule
		\multirow{2}{*}{BERT} & PointNet               & 7.49 & 12.59 & 22.25 & 31.62 & 42.63 & 31.00 & 116.58 & 6.24 & 10.37 & 19.04 & 27.91 & 39.08 & 37.00 & 102.62 \\
		& PointTrans       & 7.29 & 12.50 & 23.08 & 33.15 & 44.14 & 29.00 & 120.17 & 7.61 & 12.15 & 20.98 & 30.72 & \textbf{41.85} & \textbf{32.00} & 113.31 \\ \midrule
		\multirow{2}{*}{CLIP} & PointNet               & 7.04 & 11.87 & 21.40 & 31.01 & 42.35 & 31.00 & 113.67 & 7.57 & 11.86 & 20.97 & 30.24 & 40.82 & 33.50 & 111.46 \\
		& PointTrans       & \textbf{9.71} & \textbf{15.49} & \textbf{25.54} & \textbf{36.05} & \textbf{46.85} & \textbf{25.00} & \textbf{133.64} & \textbf{8.02} & \textbf{12.54} & \textbf{21.42} & \textbf{31.12} & 41.71 & \textbf{32.00} & \textbf{114.59} \\ \bottomrule
	\end{tabular}
	\label{table:model1}
\end{table*}

\textbf{Feature Decoder.}
The feature decoder $\psi_S$ is composed of multiple stacked cross-attention layers.
The decoder takes the masked sequence features $\text{sg}(\hat{F}^S)$ as queries, while text features $F^T$ and point features $F^P$ are alternately used as keys and values.
Specifically, in odd-numbered layers, the text features $F^T$ serve as the keys and values:
\begin{equation}
	\begin{split}
		Q = F^S W^Q, \ K &= F^T W^K, \ V = F^T W^V, \\
		\text{CrossAttn}(Q, K, &V) = \text{softmax}\left(\frac{QK^T}{\sqrt{d_k}}\right)V.
	\end{split}
	\label{eq:cross}
\end{equation}
In even-numbered layers, the point features $F^P$ act as the keys and values.
The feature decoder achieves implicit multi-modal alignment through a cross-attention reconstruction task: it takes the masked sequence features as queries and alternately uses text and point cloud features as conditional information.
Through multiple alternating iterations, the decoder forces the text and point cloud encoders to learn representations that are complementary to each other and aligned with the sequence features.
This process naturally achieves semantic alignment across the three modalities without requiring explicit positive-negative sample contrast.

\subsection{Training Objectives}
\label{sec:loss}
As shown in Fig.~\ref{fig:framework}, we employ two losses to train the architecture: a retrieval loss that explicitly enforces cross-modal alignment, and a reconstruction loss that implicitly achieves multi-modal alignment.

\textbf{Cross-modal Retrieval Loss.}
Given a mini-batch of $B$ matched text--CAD pairs $\{(\mathcal{T}_i, \mathcal{M}_i)\}_{i=1}^{B}$, we first extract the modality-specific representations $f_i^{T}$, $f_i^{S}$, and $f_i^{P}$ as described in Section~\ref{sec:pipeline}.
We adopt the bidirectional InfoNCE loss~\cite{InfoNCE} to supervise the sequence and point cloud branches independently:
\begin{equation}
	\mathcal{L}_{\text{ret}} =
	\frac{1}{2}
	\left(
	\mathcal{L}_{T \rightarrow S} +
	\mathcal{L}_{S \rightarrow T}
	\right)
	+
	\frac{1}{2}
	\left(
	\mathcal{L}_{T \rightarrow P} +
	\mathcal{L}_{P \rightarrow T}
	\right),
	\label{eq:retrieval_loss}
\end{equation}
where
\begin{equation}
	\mathcal{L}_{T \rightarrow S} =
	-\frac{1}{B}
	\sum_{i=1}^{B}
	\log
	\frac{
		\exp\bigl(\rho(f_i^{T}, f_i^{S}) / \tau\bigr)
	}{
		\sum_{j=1}^{B}
		\exp\bigl(\rho(f_i^{T}, f_j^{S}) / \tau\bigr)
	},
	\label{eq:tnc}
\end{equation}
and the remaining three terms are defined in an analogous manner.
Here, $\tau$ is a temperature parameter that controls the sharpness of the similarity distribution.

\textbf{Sequence Reconstruction Loss.}
Given the masked sequence features $\hat{F}^S_i$, along with the text and point features $F^T_i$ and $F^P_i$, the feature decoder reconstructs the complete sequence representation $\tilde{F}^S_i$ as defined in Eq.~\ref{eq:decoding}.
The reconstruction loss is defined as the mean squared error (MSE) between $\text{sg}(F^S_i)$ and $\tilde{F}^S_i$:
\begin{equation}
	\mathcal{L}_{\text{rec}} =
	\frac{1}{B}
	\sum_{i=1}^{B}
	\left\|
	\tilde{F}_i^S - \text{sg}(F_i^S)
	\right\|_2^2.
	\label{eq:recon_loss}
\end{equation}

\textbf{Overall Objective.}
The final training objective is defined as a weighted combination of the retrieval and reconstruction losses:
\begin{equation}
	\mathcal{L} =
	\mathcal{L}_{\text{ret}} +
	\lambda \, \mathcal{L}_{\text{rec}},
	\label{eq:total_loss}
\end{equation}
where $\lambda$ is a balancing coefficient.

\section{Experimental Evaluation}
\label{sec:experi}

\subsection{Experiments}

\textbf{Datasets.}
We evaluate our method on two text-to-CAD generation datasets: Text2CAD~\cite{Text2cad} and CadTranslator~\cite{Cadtranslator}.
Both datasets derive their CAD models from DeepCAD~\cite{Deepcad}, with each model accompanied by a textual description.
Accordingly, we adopt the same train/validation/test split as DeepCAD.
CAD models that cannot be converted into point cloud representations are excluded from the dataset.
After preprocessing, the final dataset comprises 119,482 training samples, 8,904 validation samples, and 8,023 test samples.
Among the four levels of descriptions in Text2CAD, we use only the Abstract Level (L0) descriptions, as they focus purely on shape semantics.
In contrast, higher-level annotations contain explicit dimensional information that is not available in practical retrieval scenarios.
Unlike the detailed descriptions in Text2CAD, CadTranslator provides simple and concise sentences for each CAD model, e.g., "an isometric view of a wall with four holes."

\textbf{Metrics.}
We evaluate text-to-CAD retrieval performance using three standard cross-modal retrieval metrics: Recall@K, Median Rank (MedR), and Rsum.
Recall@K measures the percentage of queries for which at least one correct CAD model is retrieved within the top-$K$ results.
We report Recall@1, 2, 5, 10, and 20, where higher values indicate better performance.
MedR denotes the median rank of the first correct retrieval result across all queries, with lower values being preferable.
Rsum is defined as the sum of multiple Recall@K scores, providing a comprehensive assessment of retrieval performance across different values of $K$.

\begin{table*}[t]
	\footnotesize
	\centering
	\caption{Ablation study of key model components, including the sequence encoder $\phi_S$, point encoder $\phi_P$, and feature decoder $\psi_S$. A checkmark indicates that the corresponding component and its associated modality are used.}
	\vspace{-0.1cm} 
	\begin{tabular}{ccc|ccccccc|ccccccc}
		\toprule
		\multirow{2}{*}{$\phi_S$} & \multirow{2}{*}{$\phi_P$} & \multirow{2}{*}{$\psi_S$} & \multicolumn{7}{c|}{Text2CAD}                                                                                        & \multicolumn{7}{c}{CadTranslator}              \\ \cmidrule{4-17} 
		&                           &                           & R1$\uparrow$    & R2$\uparrow$    & R5$\uparrow$    & R10$\uparrow$   & R20$\uparrow$   & MedR$\downarrow$  & Rsum$\uparrow$  & R1$\uparrow$    & R2$\uparrow$    & R5$\uparrow$    & R10$\uparrow$   & R20$\uparrow$   & MedR$\downarrow$  & Rsum$\uparrow$ \\ \midrule
		$\checkmark$              &                           &                           & 5.00          & 9.34           & 17.38          & 26.67          & 36.58          & 48.00          & 94.96           & 4.77 & 8.25 & 16.32 & 24.77 & 34.51 & 49.00 & 88.61 \\
		& $\checkmark$              &                           & 6.73          & 10.99          & 20.25          & 29.34          & 40.71          & 34.00          & 108.03          & 4.54 & 7.70 & 14.88 & 22.93 & 33.31 & 52.00 & 83.36 \\
		$\checkmark$              & $\checkmark$              &                           & 6.99          & 11.68          & 21.49          & 31.04          & 41.98          & 33.00          & 113.17          & 7.83 & 12.31 & \textbf{21.69} & 30.71 & \textbf{41.74} & \textbf{32.00} & 114.28 \\
		$\checkmark$              & $\checkmark$              & $\checkmark$              & \textbf{9.71} & \textbf{15.49} & \textbf{25.54} & \textbf{36.05} & \textbf{46.85} & \textbf{25.00} & \textbf{133.64} & \textbf{8.02} & \textbf{12.54} & 21.42 & \textbf{31.12} & 41.71 & \textbf{32.00} & \textbf{114.59} \\ \bottomrule
	\end{tabular}
	\label{table:model2}
\end{table*}

\textbf{Settings.}
For data preprocessing, the maximum length of each CAD command sequence is set to 272, following Text2CAD~\cite{Text2cad}, and 1,024 points are randomly sampled as input to the point encoder.
For the network architecture, the text encoder consists of a 4-layer Transformer encoder, while the sequence encoder is implemented using a 5-layer Transformer.
The feature decoder is constructed with 8 cross-attention layers, where half of the layers attend to textual features and the remaining half attend to point cloud features.
All text, sequence, and point features are embedded into a 256-dimensional space, resulting in a 512-dimensional representation for the final CAD and text embeddings.
The model is trained for 100 epochs using the Adam optimizer~\cite{Adam} with a learning rate of $1\times10^{-4}$.
For hyper-parameters, the masking ratio $r$ for $\hat{F}^S$ is set to 0.5 and 0.0 for Text2CAD and CadTranslator, respectively, and the loss weight $\lambda$ in Eq.~\ref{eq:total_loss} is set to 1.0, and the temperature parameter $\tau$ in Eq.~\ref{eq:retrieval_loss} is set to a learnable scalar following CLIP~\cite{CLIP}.

\begin{table}[t]
	\footnotesize
	\centering
	\caption{Performance of the proposed model on Text2CAD~\cite{Text2cad} under different masking ratios $r$ and reconstruction loss weights $\lambda$.}
	\vspace{-0.1cm} 
	\begin{tabular}{cc|ccccccc}
		\toprule
		\multicolumn{2}{c|}{\multirow{2}{*}{Parameters}} & \multicolumn{7}{c}{Text2CAD}                       \\ \cmidrule{3-9} 
		\multicolumn{2}{c|}{}                            & R1    & R2    & R5    & R10   & R20   & MedR  & Rsum  \\ \midrule
		\multirow{5}{*}{$r$}             & 0.00            & 8.94 & 14.81 & 25.44 & 35.54 & 46.63 & 25.00 & 131.35 \\
		& 0.25            & 9.57 & 15.29 & 26.19 & 36.26 & \textbf{47.24} & \textbf{24.00} & 134.55 \\
		& 0.50            & \textbf{9.71} & 15.49 & 25.54 & 36.05 & 46.85 & 25.00 & 133.64 \\
		& 0.75            & 9.67 & \textbf{15.83} & \textbf{26.36} & 36.51 & 47.03 & 25.00 & \textbf{135.40} \\
		& 1.00            & 9.57 & 15.24 & 26.08 & \textbf{36.83} & 46.83 & 25.00 & 134.10 \\ \midrule \midrule
		\multirow{4}{*}{$\lambda$}             & 0.1             & \textbf{9.81} & 15.41 & 25.91 & 36.28 & 46.95 & 25.00 & 134.26 \\
		& 0.5             & 9.57 & \textbf{15.54} & \textbf{26.21} & \textbf{36.58} & \textbf{47.11} & \textbf{24.00} & \textbf{135.02} \\
		& 1.0             & 9.71 & 15.49 & 25.54 & 36.05 & 46.85 & 25.00 & 133.64 \\
		& 2.0             & 9.22 & 15.24 & 25.73 & 35.95 & 46.63 & 25.00 & 132.77 \\ \bottomrule
	\end{tabular}
	\label{table:params}
\end{table}

\begin{table}[t]
	\footnotesize
	\centering
	\caption{Performance of the proposed model on CadTranslator~\cite{Cadtranslator} under different masking ratios $r$ and reconstruction loss weights $\lambda$.}
	\vspace{-0.1cm} 
	\begin{tabular}{cc|ccccccc}
		\toprule
		\multicolumn{2}{c|}{\multirow{2}{*}{Parameters}} & \multicolumn{7}{c}{CadTranslator}                       \\ \cmidrule{3-9} 
		\multicolumn{2}{c|}{}                            & R1    & R2    & R5    & R10   & R20   & MedR  & Rsum  \\ \midrule
		\multirow{5}{*}{$r$}             & 0.00            & \textbf{8.02} & \textbf{12.54} & \textbf{21.42} & \textbf{31.12} & \textbf{41.71} & \textbf{32.00} & \textbf{114.59} \\
		& 0.25            & 7.59 & 12.07 & 21.13 & 30.51 & 41.21 & 33.00 & 112.51 \\
		& 0.50            & 6.38 & 10.39 & 18.65 & 27.28 & 38.37 & 39.00 & 101.08 \\
		& 0.75            & 6.06 & 10.11 & 18.32 & 27.44 & 38.20 & 38.00 & 100.19 \\
		& 1.00            & 7.33 & 11.99 & 21.12 & 30.39 & 41.41 & 33.00 & 112.23 \\ \midrule \midrule
		\multirow{4}{*}{$\lambda$}             & 0.1             & 6.85 & 11.18 & 20.09 & 29.54 & 40.50 & 34.00 & 108.15 \\
		& 0.5             & 6.43 & 10.74 & 19.60 & 28.78 & 39.89 & 35.00 & 105.44 \\
		& 1.0             & \textbf{8.02} & \textbf{12.54} & \textbf{21.42} & \textbf{31.12} & \textbf{41.71} & \textbf{32.00} & \textbf{114.59} \\
		& 2.0             & 7.65 & 12.00 & 20.86 & 30.50 & 41.17 & 33.00 & 112.18 \\ \bottomrule
	\end{tabular}
	\label{table:params2}
\end{table}

In the following sections, we first present quantitative results demonstrating the effectiveness of the model components and the exploration of hyperparameters. We then provide a qualitative analysis to verify the geometric sensitivity and superiority of our method. Finally, we compare the proposed model with existing state-of-the-art (SOTA) approaches.

\subsection{Quantitative Results}

\textbf{Verification of model components.}
We first investigate the impact of different text tokenizers for $\phi_T$ and different backbone architectures for $\phi_P$.
Specifically, we compare BERT~\cite{BERT} and CLIP~\cite{CLIP} as text tokenizers, and implement PointNet~\cite{PointNet} and PointTransformer~\cite{PointTransformer1,PointTransformer2} as point encoders.
The quantitative results are reported in Table~\ref{table:model1}.
As shown, CLIP-based tokenization consistently outperforms BERT, while the PointTransformer-based encoder achieves better performance than PointNet.
Specifically, using CLIP with PointTransformer yields a significant improvement over BERT with PointNet: R1 accuracy increases from 7.49 to 9.71 on Text2CAD and from 6.24 to 8.02 on CadTranslator.
Accordingly, we adopt CLIP for textual encoding and PointTransformer for point encoding.

Next, we examine the contribution of multi-modal CAD representations and the proposed feature decoder.
To this end, we implement multiple frameworks for text-CAD matching: encoding sequence only with $\phi_S$, encoding point clouds only with $\phi_P$, encoding both sequence and point clouds with $\phi_S$ and $\phi_P$, using $\phi_S$ and $\phi_P$ for sequence and point encoding meanwhile using $\psi_S$ for feature reconstruction.
The results on Text2CAD and CadTranslator are reported in Table~\ref{table:model2}.
From this table, we draw the following conclusions. First, using both procedural sequences and geometric point clouds significantly outperforms single-modality variants. For instance, R1 accuracy improves from 4.77 (sequence only) and 4.54 (point only) to 7.83 (both). Second, introducing the feature decoder leads to a substantial performance gain, confirming its effectiveness in facilitating implicit multi-modal alignment. Notably, the improvement from the feature decoder is larger on Text2CAD (R1 from 6.99 to 9.71) than on CadTranslator (R1 from 7.83 to 8.02).

\begin{figure}[t]
	\centering
	\begin{subfigure}{0.24\textwidth}
		\centering
		\includegraphics[width=\linewidth]{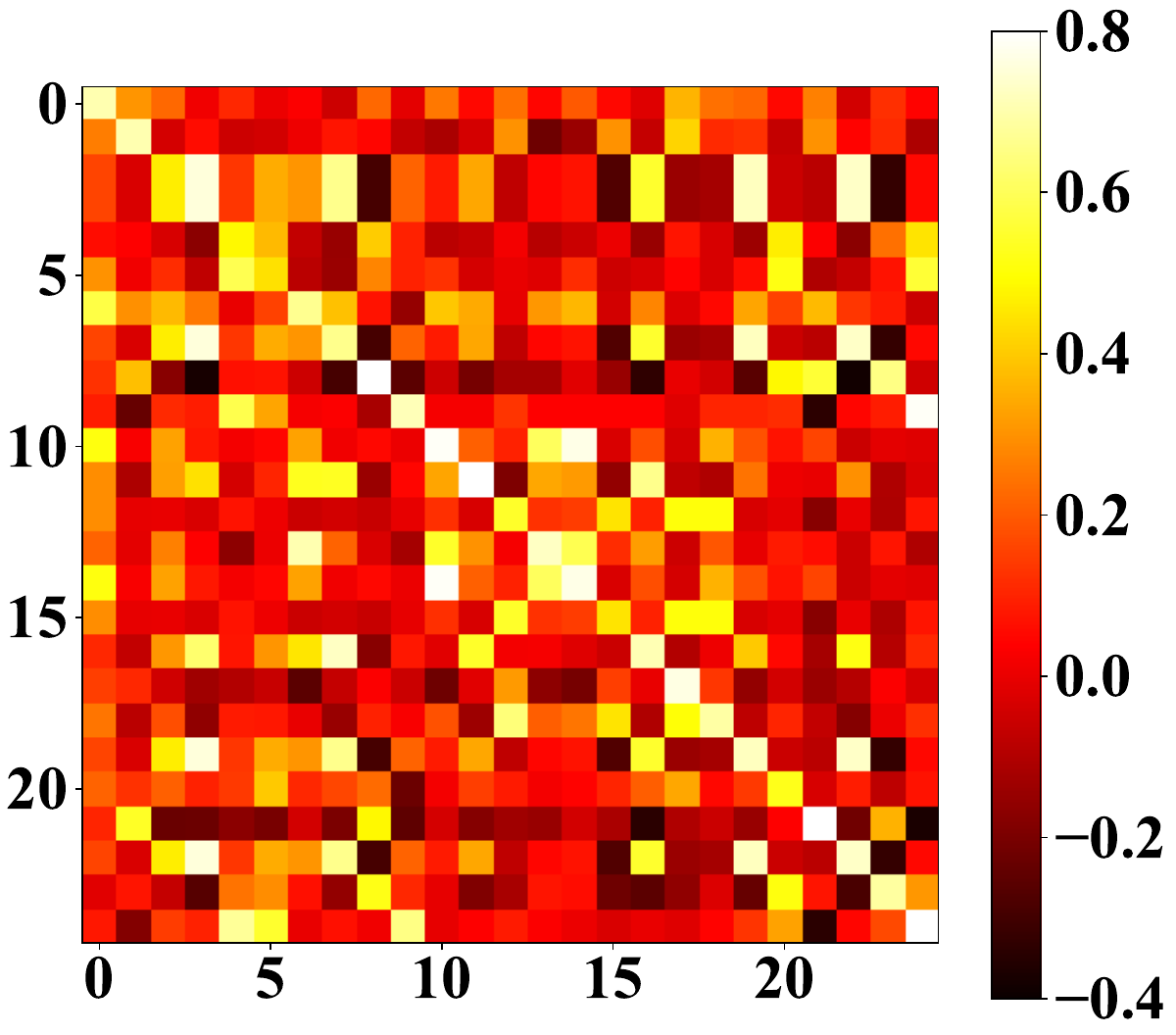}
		\vspace{-6mm}
		\caption{Sequence-only model}

		\label{fig:heatmap1}
	\end{subfigure}
	\begin{subfigure}{0.24\textwidth}
		\centering
		\includegraphics[width=\linewidth]{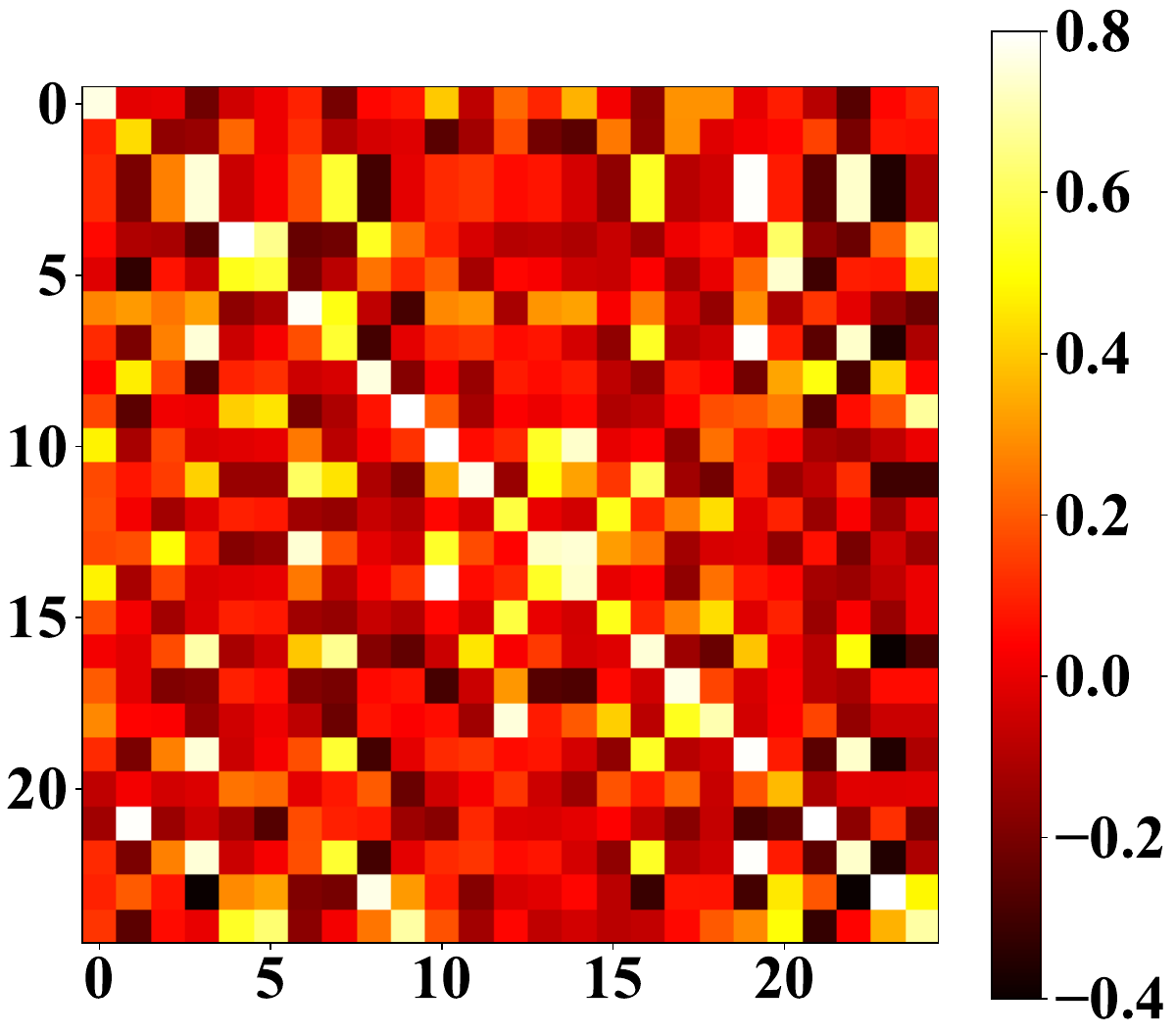}
		\vspace{-6mm}
		\caption{Our approach}
		\label{fig:heatmap2}
	\end{subfigure}
	\vspace{-6mm}
	\caption{Visualization of similarity score. The diagonal elements represent the similarity scores for positive sample pairs, while the remaining elements represent the similarity scores for negative sample pairs.
	}
	\label{fig:heatmap}
\end{figure}

\begin{figure*}
	\includegraphics[width=0.99 \textwidth]{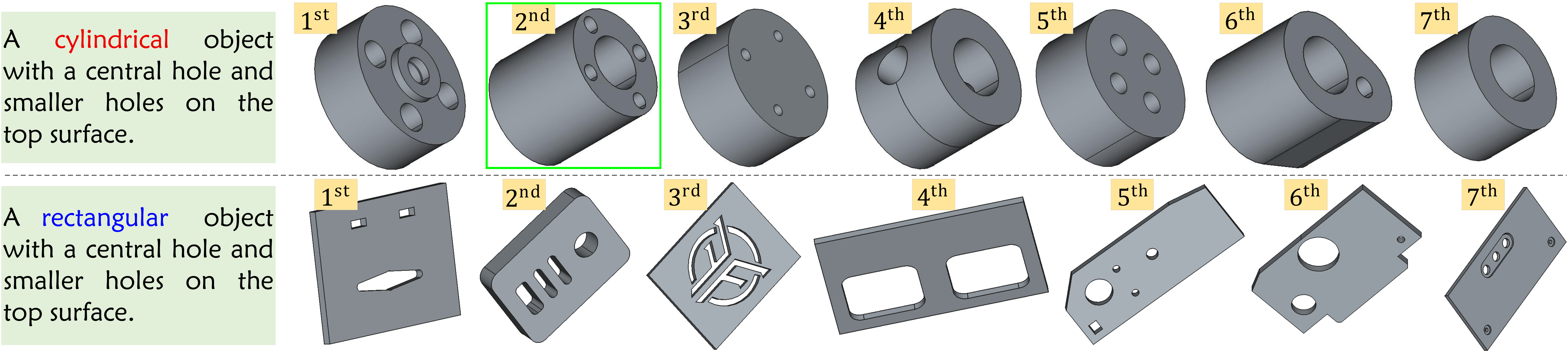}
	\vspace{-0.1cm} 
	\caption{
		Exploration of geometry sensitivity.
		First row: the original textual query and its retrieval results, where the second retrieved CAD model (highlighted by a green box) is the correct match.
		Second row: the word ``\textcolor{red}{\texttt{cylindrical}}'' in the original query is replaced with ``\textcolor{blue}{\texttt{rectangular}}'', and the corresponding retrieval results reveal the change in geometric preference.
	}
	\label{fig:geo}
\end{figure*}

\begin{figure*}
	\includegraphics[width=0.99 \textwidth]{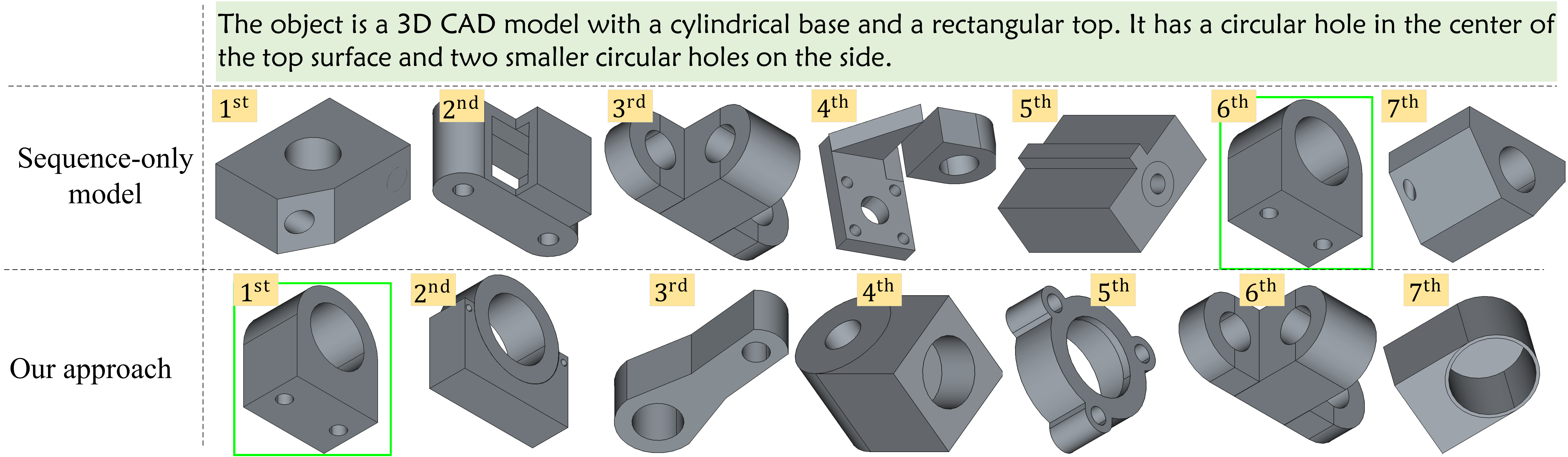}
	\vspace{-0.1cm} 
	\caption{
		Visualization of retrieval results by sequence model and our approach, where the CAD model highlighted by a green box denotes the positive sample.
	}
	\label{fig:rank}
\end{figure*}

Finally, we compare the similarity scores learned by the sequence-only model (i.e., encoding sequences solely with $\phi_S$) and those learned by our full approach in Fig.~\ref{fig:heatmap}. To this end, we randomly select 25 text-CAD pairs from the Text2CAD~\cite{Text2cad} test set and compute a $25 \times 25$ similarity matrix for each model. In this matrix, diagonal entries correspond to the similarity scores of positive (matching) pairs, while off-diagonal entries correspond to negative (non-matching) pairs. The ground-truth similarity matrix is an identity matrix $I \in \mathbb{R}^{25 \times 25}$.
As shown in Fig.~\ref{fig:heatmap}, our approach learns more discriminative representations: off-diagonal entries exhibit darker red hues, indicating lower similarity scores for negative sample pairs.

\textbf{Exploration of model hyperparameters.}
We further study the influence of two hyperparameters: the masking ratio $r$ for the feature decoder and the reconstruction loss weight $\lambda$ in Eq.~\ref{eq:total_loss}.
The masking ratio $r$ controls the proportion of masked sequence features. When $r = 0$, no masking is applied to the sequence features; when $r = 1$, the sequence features are recovered entirely from an all-zero vector. The loss weight $\lambda$ balances the contribution between the retrieval loss and the reconstruction loss.
Results on Text2CAD and CadTranslator are reported in Table~\ref{table:params} and Table~\ref{table:params2}, respectively.

For the masking ratio $r$, on Text2CAD, varying $r$ from 0.00 to 1.00 yields relatively stable performance, with R1 fluctuating between 8.94 and 9.71. The best R1 (9.71) is achieved at $r=0.50$, while the highest Rsum (135.40) is obtained at $r=0.75$. Median rank remains largely unchanged (24–25). On CadTranslator, the impact of $r$ is more noticeable. The optimal performance occurs at $r=0.00$ (R1=8.02, Rsum=114.59), and increasing $r$ generally degrades results, especially at $r=0.75$ where R1 drops to 6.06. This suggests that aggressive masking of sequence features may be less beneficial when the geometric descriptions are simpler, as in CadTranslator.

For the reconstruction weight $\lambda$, on Text2CAD, the model achieves competitive results across $\lambda \in \{0.1,0.5,1.0,2.0\}$. The best R1 (9.81) is observed at $\lambda=0.1$, while the highest Rsum (135.02) and best MedR (24) are obtained at $\lambda=0.5$. The default $\lambda=1.0$ also yields strong performance (R1=9.71). On CadTranslator, the optimal $\lambda$ is clearly $1.0$, which achieves the best R1 (8.02), Rsum (114.59), and MedR (32). Increasing $\lambda$ to 2.0 or decreasing to 0.5 leads to moderate performance drops. Overall, the model is reasonably robust to $\lambda$ on Text2CAD, whereas $\lambda=1.0$ is recommended for CadTranslator.

\begin{table*}[t]
	\footnotesize
	\centering
	\caption{Comparison with SOTA.}
	\vspace{-0.1cm} 
\begin{tabular}{l|ccccccc|ccccccc}
	\toprule
	\multirow{2}{*}{Method}                      & \multicolumn{7}{c|}{Text2CAD}                                                                                          & \multicolumn{7}{c}{CadTranslator}                                                                              \\ \cmidrule{2-15} 
	& R1$\uparrow$  & R2$\uparrow$   & R5$\uparrow$   & R10$\uparrow$  & R20$\uparrow$  & MedR$\downarrow$ & Rsum$\uparrow$  & R1$\uparrow$ & R2$\uparrow$ & R5$\uparrow$ & R10$\uparrow$ & R20$\uparrow$ & MedR$\downarrow$ & Rsum$\uparrow$ \\ \midrule
	Parts2Words~\cite{parts2words} & 4.81          & 8.43           & 15.73          & 23.60          & 32.98          & 61.00            & 88.97           & 4.16         & 7.35         & 13.60        & 21.10         & 30.32         & 64.00           & 76.55          \\
	TriColo~\cite{Tricolo} & 2.12          & 3.98           & 8.53          & 13.60          & 20.47          & 144.00            & 48.69           & 2.77         & 4.92         & 9.57        & 15.40         & 22.44         & 123.00           & 55.05          \\
	SCA3D~\cite{sca3d}     & 6.33          & 10.81           & 19.27           & 27.50           & 37.17           & 45.00             & 103.04            & 5.58         & 9.25         & 16.33         & 23.70          & 33.15          & 58.00             & 87.97           \\
	RoMa~\cite{roma}       & 3.80          & 6.66           & 12.23           & 18.60           & 26.74           & 99.00             & 67.97            & 3.84         & 6.42         & 11.78         & 18.40          & 25.71          & 110.00             & 66.11           \\ \midrule
	IMMA (Ours)                                  & \textbf{9.71} & \textbf{15.49} & \textbf{25.54} & \textbf{36.05} & \textbf{46.85} & \textbf{25.00}   & \textbf{133.64} & \textbf{8.02}         & \textbf{12.54}        & \textbf{21.42}        & \textbf{31.12}         & \textbf{41.71}         & \textbf{32.00}            & \textbf{114.59}         \\ \bottomrule
\end{tabular}
	\label{table:sota}
\end{table*}

\begin{table*}[t]
	\footnotesize
	\centering
	\caption{Comparison with other fusion strategies.}
	\vspace{-0.1cm} 
	\begin{tabular}{l|ccccccc|ccccccc}
		\toprule
		\multirow{2}{*}{Fusion strategy} & \multicolumn{7}{c|}{Text2CAD}                  & \multicolumn{7}{c}{CadTranslator}              \\ \cmidrule{2-15} 
		& R1$\uparrow$    & R2$\uparrow$    & R5$\uparrow$    & R10$\uparrow$   & R20$\uparrow$   & MedR$\downarrow$  & Rsum$\uparrow$  & R1$\uparrow$    & R2$\uparrow$    & R5$\uparrow$    & R10$\uparrow$   & R20$\uparrow$   & MedR$\downarrow$  & Rsum$\uparrow$   \\ \midrule
		Concat                           & 6.99          & 11.68          & 21.49          & 31.04          & 41.98          & 33.00          & 113.17  & 7.83 & 12.31 & \textbf{21.69} & 30.71 & \textbf{41.74} & \textbf{32.00} & 114.28 \\
		Concat \& Linear                    & 6.14 & 10.86 & 19.67 & 28.42 & 39.15 & 38.00 & 104.24 & 7.02 & 11.31 & 19.46 & 28.26 & 38.78 & 39.00 & 104.84 \\
		Concat \& SelfAttn                 & 6.27 & 10.73 & 20.24 & 29.27 & 39.89 & 37.00 & 106.39 & 5.70 & 9.70 & 18.21 & 27.46 & 38.56 & 37.50 & 99.63 \\
		CrossAttn                         & 6.69 & 11.16 & 20.73 & 29.90 & 40.70 & 36.00 & 109.17 & 4.64 & 8.50 & 16.76 & 26.03 & 37.46 & 39.00 & 93.39 \\
		Modulation                       & 6.56 & 11.48 & 20.30 & 29.74 & 40.86 & 35.00 & 109.14 & 7.45 & 11.86 & 20.61 & 29.40 & 39.56 & 38.00 & 108.87 \\ \midrule
		IMMA (Ours)                             & \textbf{9.71} & \textbf{15.49} & \textbf{25.54} & \textbf{36.05} & \textbf{46.85} & \textbf{25.00} & \textbf{133.64} & \textbf{8.02} & \textbf{12.54} & 21.42 & \textbf{31.12} & 41.71 & \textbf{32.00} & \textbf{114.59} \\ \bottomrule
	\end{tabular}
	\label{table:fusion}
\end{table*}

\subsection{Qualitative Analysis}
In this subsection, we demonstrate the effectiveness and geometric sensitivity of the proposed model through visualizations. First, we conduct a qualitative analysis to examine the geometric sensitivity of our method. Specifically, we modify geometric attributes in the textual queries and observe the corresponding retrieval results. As illustrated in Fig.~\ref{fig:geo}, altering a single geometric descriptor in the query (from ``\textcolor{red}{\texttt{cylindrical}}'' to ``\textcolor{blue}{\texttt{rectangular}}'') leads to semantically consistent changes in the retrieved CAD models, demonstrating that our model can identify geometry-consistent CAD models from given text descriptions.

Next, we compare the sequence-only model with our full approach by visualizing the retrieval results. Fig.~\ref{fig:rank} presents the retrieval ranks for a given sentence. The sequence-only model fails to locate CAD models relevant to the text description. In contrast, our approach successfully retrieves semantically matching CAD models: the top-ranked results include a cylindrical base and a rectangular top with holes.

\subsection{Comparisons}
In this subsection, we compare our approach, Implicit Multi-Modal Alignment (IMMA), with existing text-to-3D shape retrieval models and other multi-modal fusion strategies.

First, we present a comparison with state-of-the-art (SOTA) methods in Table~\ref{table:sota}, including Parts2Words~\cite{parts2words} (CVPR'23), TriColo~\cite{Tricolo} (WACV'24), SCA3D~\cite{sca3d} (ICRA'25), and RoMa~\cite{roma} (TMM'25). The following adaptations were made to suit our task.

For Parts2Words~\cite{parts2words}, segmentation supervision was disabled and a retrieval‑only training mode was adopted. Unlike SCA3D, Parts2Words retains its pseudo‑part construction pipeline, where pseudo‑part labels are generated from the maximum‑response categories of semantic predictions and then aggregated via point‑to‑part pooling. The first‑stage training epoch was set to 0, enabling retrieval optimization from the very first epoch.
For TriCoLo~\cite{Tricolo}, the main adaptation concerns the input layer. Since our 3D input consists of 1024‑dimensional geometric features rather than the original 3‑channel voxel representation, the input channel number of the first layer in the voxel branch was changed from 3 to 1024, while the rest of the architecture remained unchanged.
For SCA3D~\cite{sca3d}, because reliable part‑level segmentation annotations are unavailable in our setting, semantic segmentation supervision was disabled. Under the $K=1$ configuration, its part construction was modified to single‑part global average pooling. Additionally, the EMD cost was changed from the original ReLU‑truncated dot‑product form to a non‑negative truncated distance based on feature dot products, making it more suitable for retrieval training without segmentation supervision.
For RoMa~\cite{roma}, we adopted its BiGRU text encoding branch. The main architecture was kept intact, and we only supplemented the evaluation with Text2CAD‑style text‑to‑shape retrieval metrics to ensure comparability under a unified protocol.

As shown in Table~\ref{table:sota}, our model substantially outperforms all existing SOTA methods. For instance, among the existing approaches, SCA3D~\cite{sca3d} achieves the best performance, with R1 scores of 6.33 and 5.58 on Text2CAD and CadTranslator respectively, which are far below those of IMMA (9.71 and 8.02). These results strongly demonstrate the effectiveness of our model.

Next, considering that most existing works~\cite{TEFAL,Eclipse} fuse multi‑modal data into a final representation, we compare our approach (concatenation with implicit alignment) against several multi‑modal fusion methods: Concat \& Linear, Concat \& SelfAttn, CrossAttn, and Modulation. Specifically, Concat \& Linear applies linear layers to the concatenated sequence and point features; Concat \& SelfAttn applies self‑attention layers to the concatenated features; CrossAttn performs cross‑attention between sequence and point features; and Modulation employs modulation layers to fuse the two modalities. The results are reported in Table~\ref{table:fusion}.
We observe that the multi‑modal fusion strategies are inferior to both the simple concatenation baseline and our method. This is likely due to the significant modality discrepancy between command sequences and point clouds. In other domains, such as text‑to‑video retrieval, fusing video frames with audio is effective because video and audio are temporally aligned and can play complementary roles.

\section{Conclusions}
\label{sec:conclu}
In this paper, we introduced \emph{text-to-CAD retrieval} as a new cross-modal retrieval task, addressing the semantic access to large-scale CAD repositories using natural language queries.
By leveraging paired text-CAD annotations from existing generation datasets, this task provides a practical and scalable paradigm for CAD model reuse in industrial workflows.
We further presented a strong baseline framework that learns multi-modal CAD representations from both procedural sequences and geometric point clouds, and employed an auxiliary feature decoder to encourage implicit cross-modal alignment during training.
Experimental results demonstrated the effectiveness and geometric sensitivity of the proposed approach.
We hope this work can stimulate further research on text-based CAD understanding and serve as a foundation of downstream retrieval-augmented CAD generation.

\bibliographystyle{IEEEtran}
\bibliography{sample-base}

@STRING{CVPR = "IEEE Conference on Computer Vision and Pattern Recognition"}

@STRING{ICCV = " IEEE International Conference on Computer Vision"}

@STRING{ECCV = "In Proceedings of the European Conference on Computer Vision"}

@STRING{NeurIPS = "Conference on Neural Information Processing Systems"}

@STRING{ICLR = "International Conference on Learning Representations"}

@STRING{ICML = "International Conference on Machine Learning"}

@STRING{AAAI = "Proceedings of the AAAI Conference on Artificial Intelligence"}

@inproceedings{Deepcad,
	title={Deepcad: A deep generative network for computer-aided design models},
	author={Wu, Rundi and Xiao, Chang and Zheng, Changxi},
	booktitle=ICCV,
	pages={6772--6782},
	year={2021}
}

@article{Complexgen,
	title={Complexgen: Cad reconstruction by b-rep chain complex generation},
	author={Guo, Haoxiang and Liu, Shilin and Pan, Hao and Liu, Yang and Tong, Xin and Guo, Baining},
	journal={ACM Transactions on Graphics},
	volume={41},
	number={4},
	pages={1--18},
	year={2022},
}

@article{Brepgen,
	title={Brepgen: A b-rep generative diffusion model with structured latent geometry},
	author={Xu, Xiang and Lambourne, Joseph and Jayaraman, Pradeep and Wang, Zhengqing and Willis, Karl and Furukawa, Yasutaka},
	journal={ACM Transactions on Graphics},
	volume={43},
	number={4},
	pages={1--14},
	year={2024},
}

@inproceedings{SkexGen,
	title={SkexGen: Autoregressive Generation of CAD Construction Sequences with Disentangled Codebooks},
	author={Xu, Xiang and Willis, Karl DD and Lambourne, Joseph G and Cheng, Chin-Yi and Jayaraman, Pradeep Kumar and Furukawa, Yasutaka},
	booktitle=ICML,
	pages={24698--24724},
	year={2022},
}

@inproceedings{HNC-CAD,
	title={Hierarchical Neural Coding for Controllable CAD Model Generation},
	author={Xu, Xiang and Jayaraman, Pradeep Kumar and Lambourne, Joseph George and Willis, Karl DD and Furukawa, Yasutaka},
	booktitle={International Conference on Machine Learning},
	pages={38443--38461},
	year={2023},
	organization={PMLR}
}

@article{Diffusion-CAD,
	title={Diffusion-cad: Controllable diffusion model for generating computer-aided design models},
	author={Zhang, Aijia and Jia, Weiqiang and Zou, Qiang and Feng, Yixiong and Wei, Xiaoxiang and Zhang, Ye},
	journal={IEEE Transactions on Visualization and Computer Graphics},
	year={2025},
}

@inproceedings{CAD-Llama,
	title={CAD-Llama: leveraging large language models for computer-aided design parametric 3D model generation},
	author={Li, Jiahao and Ma, Weijian and Li, Xueyang and Lou, Yunzhong and Zhou, Guichun and Zhou, Xiangdong},
	booktitle=CVPR,
	pages={18563--18573},
	year={2025}
}

@inproceedings{CMT,
	title={Cmt: A cascade mar with topology predictor for multimodal conditional cad generation},
	author={Wu, Jianyu and Wang, Yizhou and Yue, Xiangyu and Ma, Xinzhu and Guo, Jinyang and Zhou, Dongzhan and Ouyang, Wanli and Tang, Shixiang},
	booktitle=ICCV,
	pages={7014--7024},
	year={2025}
}

@inproceedings{RECAD,
	title={Revisiting cad model generation by learning raster sketch},
	author={Li, Pu and Zhang, Wenhao and Guo, Jianwei and Chen, Jinglu and Yan, Dong-Ming},
	booktitle={Proceedings of the AAAI Conference on Artificial Intelligence},
	volume={39},
	number={5},
	pages={4869--4877},
	year={2025}
}

@article{Text2cad,
	title={Text2cad: Generating sequential cad designs from beginner-to-expert level text prompts},
	author={Khan, Mohammad S and Sinha, Sankalp and Sheikh, Talha U and Stricker, Didier and Ali, Sk A and Afzal, Muhammad Z},
	journal={Advances in Neural Information Processing Systems},
	volume={37},
	pages={7552--7579},
	year={2024}
}

@inproceedings{Cadtranslator,
	title={Cad translator: An effective drive for text to 3d parametric computer-aided design generative modeling},
	author={Li, Xueyang and Song, Yu and Lou, Yunzhong and Zhou, Xiangdong},
	booktitle={Proceedings of the 32nd ACM International Conference on Multimedia},
	pages={8461--8470},
	year={2024}
}

@inproceedings{FlexCAD,
	title={FlexCAD: Unified and Versatile Controllable CAD Generation with Fine-tuned Large Language Models},
	author={Zhang, Zhanwei and Sun, Shizhao and Wang, Wenxiao and Cai, Deng and Bian, Jiang},
	booktitle=ICLR,
	year={2025}
}

@article{Transformer,
	title={Attention is all you need},
	author={Vaswani, Ashish and Shazeer, Noam and Parmar, Niki and Uszkoreit, Jakob and Jones, Llion and Gomez, Aidan N and Kaiser, {\L}ukasz and Polosukhin, Illia},
	journal={Advances in neural information processing systems},
	volume={30},
	year={2017}
}

@inproceedings{PointNet,
	title={Pointnet: Deep learning on point sets for 3d classification and segmentation},
	author={Qi, Charles R and Su, Hao and Mo, Kaichun and Guibas, Leonidas J},
	booktitle=CVPR,
	pages={652--660},
	year={2017}
}

@article{Adam,
	title={Adam: A method for stochastic optimization},
	author={Kingma, Diederik P and Ba, Jimmy},
	journal={arXiv preprint arXiv:1412.6980},
	year={2014}
}

@article{t2i,
	title={Towards unified bijective image-text generation for text-to-image person re-identification},
	author={Wang, Qianqian and Ma, Xiaoguang and Jiang, Xiaoyu and Ji, Jianmin and Pan, Honghu},
	journal={Knowledge-Based Systems},
	pages={114014},
	year={2025},
	publisher={Elsevier}
}

@inproceedings{Eclipse,
	title={Eclipse: Efficient long-range video retrieval using sight and sound},
	author={Lin, Yan-Bo and Lei, Jie and Bansal, Mohit and Bertasius, Gedas},
	booktitle=ECCV,
	pages={413--430},
	year={2022}
}

@inproceedings{CLIP,
	title={Learning transferable visual models from natural language supervision},
	author={Radford, Alec and Kim, Jong Wook and Hallacy, Chris and Ramesh, Aditya and Goh, Gabriel and Agarwal, Sandhini and Sastry, Girish and Askell, Amanda and Mishkin, Pamela and Clark, Jack and others},
	booktitle=ICML,
	pages={8748--8763},
	year={2021},
	organization={PMLR}
}

@inproceedings{BERT,
	title={Bert: Pre-training of deep bidirectional transformers for language understanding},
	author={Devlin, Jacob and Chang, Ming-Wei and Lee, Kenton and Toutanova, Kristina},
	booktitle={Proceedings of the 2019 conference of the North American chapter of the association for computational linguistics: human language technologies, volume 1 (long and short papers)},
	pages={4171--4186},
	year={2019}
}

@inproceedings{PointTransformer1,
	title={Point transformer},
	author={Zhao, Hengshuang and Jiang, Li and Jia, Jiaya and Torr, Philip HS and Koltun, Vladlen},
	booktitle={Proceedings of the IEEE/CVF international conference on computer vision},
	pages={16259--16268},
	year={2021}
}

@article{PointTransformer2,
	title={Point transformer v2: Grouped vector attention and partition-based pooling},
	author={Wu, Xiaoyang and Lao, Yixing and Jiang, Li and Liu, Xihui and Zhao, Hengshuang},
	journal={Advances in Neural Information Processing Systems},
	volume={35},
	pages={33330--33342},
	year={2022}
}

@article{InfoNCE,
	title={Representation learning with contrastive predictive coding},
	author={Oord, Aaron van den and Li, Yazhe and Vinyals, Oriol},
	journal={arXiv preprint arXiv:1807.03748},
	year={2018}
}

@inproceedings{RAG1,
	title={Evaluating retrieval quality in retrieval-augmented generation},
	author={Salemi, Alireza and Zamani, Hamed},
	booktitle={Proceedings of the 47th International ACM SIGIR Conference on Research and Development in Information Retrieval},
	pages={2395--2400},
	year={2024}
}

@article{RAG2,
	title={Retrieval-augmented generation for knowledge-intensive nlp tasks},
	author={Lewis, Patrick and Perez, Ethan and Piktus, Aleksandra and Petroni, Fabio and Karpukhin, Vladimir and Goyal, Naman and K{\"u}ttler, Heinrich and Lewis, Mike and Yih, Wen-tau and Rockt{\"a}schel, Tim and others},
	journal={Advances in neural information processing systems},
	volume={33},
	pages={9459--9474},
	year={2020}
}

@article{RAG3,
	title={Retrieval-augmented generation for large language models: A survey},
	author={Gao, Yunfan and Xiong, Yun and Gao, Xinyu and Jia, Kangxiang and Pan, Jinliu and Bi, Yuxi and Dai, Yixin and Sun, Jiawei and Wang, Haofen and Wang, Haofen},
	journal={arXiv preprint arXiv:2312.10997},
	volume={2},
	number={1},
	year={2023}
}

@article{LLaMA3,
	title={The llama 3 herd of models},
	author={Dubey, Abhimanyu and Jauhri, Abhinav and Pandey, Abhinav and Kadian, Abhishek and Al-Dahle, Ahmad and Letman, Aiesha and Mathur, Akhil and Schelten, Alan and Yang, Amy and Fan, Angela and others},
	journal={arXiv e-prints},
	pages={arXiv--2407},
	year={2024}
}

@article{Gpt-4,
	title={Gpt-4 technical report},
	author={Achiam, Josh and Adler, Steven and Agarwal, Sandhini and Ahmad, Lama and Akkaya, Ilge and Aleman, Florencia Leoni and Almeida, Diogo and Altenschmidt, Janko and Altman, Sam and Anadkat, Shyamal and others},
	journal={arXiv preprint arXiv:2303.08774},
	year={2023}
}

@article{VQ-VAE,
	title={Neural discrete representation learning},
	author={Van Den Oord, Aaron and Vinyals, Oriol and others},
	journal=NeurIPS,
	volume={30},
	year={2017}
}

@inproceedings{text2shape,
	title={Text2shape: Generating shapes from natural language by learning joint embeddings},
	author={Chen, Kevin and Choy, Christopher B and Savva, Manolis and Chang, Angel X and Funkhouser, Thomas and Savarese, Silvio},
	booktitle={Asian Conference on Computer Vision},
	pages={100--116},
	year={2018},
	organization={Springer}
}

@inproceedings{Y2Seq2Seq,
	title={Y2Seq2Seq: Cross-modal representation learning for 3D shape and text by joint reconstruction and prediction of view and word sequences},
	author={Han, Zhizhong and Shang, Mingyang and Wang, Xiyang and Liu, Yu-Shen and Zwicker, Matthias},
	booktitle=AAAI,
	volume={33},
	number={01},
	pages={126--133},
	year={2019}
}

@inproceedings{parts2words,
	title={Parts2words: Learning joint embedding of point clouds and texts by bidirectional matching between parts and words},
	author={Tang, Chuan and Yang, Xi and Wu, Bojian and Han, Zhizhong and Chang, Yi},
	booktitle=CVPR,
	pages={6884--6893},
	year={2023}
}

@inproceedings{Tricolo,
	title={Tricolo: Trimodal contrastive loss for text to shape retrieval},
	author={Ruan, Yue and Lee, Han-Hung and Zhang, Yiming and Zhang, Ke and Chang, Angel X},
	booktitle={Proceedings of the IEEE/CVF Winter Conference on Applications of Computer Vision},
	pages={5815--5825},
	year={2024}
}

@inproceedings{sca3d,
	title={Sca3d: Enhancing cross-modal 3d retrieval via 3d shape and caption paired data augmentation},
	author={Ren, Junlong and Wu, Hao and Xiong, Hui and Wang, Hao},
	booktitle={2025 IEEE International Conference on Robotics and Automation (ICRA)},
	pages={9550--9557},
	year={2025},
	organization={IEEE}
}

@article{roma,
	title={Pointcloud-text matching: Benchmark dataset and baseline},
	author={Feng, Yanglin and Qin, Yang and Peng, Dezhong and Zhu, Hongyuan and Peng, Xi and Hu, Peng},
	journal={IEEE Transactions on Multimedia},
	year={2025},
	publisher={IEEE}
}

@inproceedings{TEFAL,
	title={Audio-enhanced text-to-video retrieval using text-conditioned feature alignment},
	author={Ibrahimi, Sarah and Sun, Xiaohang and Wang, Pichao and Garg, Amanmeet and Sanan, Ashutosh and Omar, Mohamed},
	booktitle=ICCV,
	pages={12020--12030},
	year={2023}
}

@article{ACD,
	title={Unified conditional image generation for visible-infrared person re-identification},
	author={Pan, Honghu and Pei, Wenjie and Li, Xin and He, Zhenyu},
	journal={IEEE Transactions on Information Forensics and Security},
	volume={19},
	pages={9026--9038},
	year={2024},
	publisher={IEEE}
}

@inproceedings{LAVIMO,
	title={Tri-modal motion retrieval by learning a joint embedding space},
	author={Yin, Kangning and Zou, Shihao and Ge, Yuxuan and Tian, Zheng},
	booktitle=CVPR,
	pages={1596--1605},
	year={2024}
}

@inproceedings{TriModal,
	title={Enhanced cross-modal 3d retrieval via tri-modal reconstruction},
	author={Ren, Junlong and Wang, Hao},
	booktitle={2025 IEEE International Conference on Multimedia and Expo (ICME)},
	pages={1--6},
	year={2025},
	organization={IEEE}
}

@article{CoCa,
	title={CoCa: Contrastive Captioners are Image-Text Foundation Models},
	author={Yu, Jiahui and Wang, Zirui and Vasudevan, Vijay and Yeung, Legg and Seyedhosseini, Mojtaba and Wu, Yonghui},
	journal={Transactions on Machine Learning Research}
}

@article{Mixtral,
	title={Mixtral of experts},
	author={Jiang, Albert Q and Sablayrolles, Alexandre and Roux, Antoine and Mensch, Arthur and Savary, Blanche and Bamford, Chris and Chaplot, Devendra Singh and Casas, Diego de las and Hanna, Emma Bou and Bressand, Florian and others},
	journal={arXiv preprint arXiv:2401.04088},
	year={2024}
}

@misc{Llavanext,
	title={Llavanext: Improved reasoning, ocr, and world knowledge},
	author={Liu, Haotian and Li, Chunyuan and Li, Yuheng and Li, Bo and Zhang, Yuanhan and Shen, Sheng and Lee, Yong Jae},
	year={2024}
}

@article{tii1,
	title={Img2cad: Conditioned 3-d cad model generation from single image with structured visual geometry},
	author={Chen, Tianrun and Yu, Chunan and Hu, Yuanqi and Li, Jing and Xu, Tao and Cao, Runlong and Zhu, Lanyun and Zang, Ying and Zhang, Yong and Li, Zejian and others},
	journal={IEEE Transactions on Industrial Informatics},
	year={2025},
	publisher={IEEE}
}

@article{tii2,
	title={View-based 3-D CAD model retrieval with deep residual networks},
	author={Zhang, Chao and Zhou, Guanghui and Yang, Haidong and Xiao, Zhongdong and Yang, Xiongjun},
	journal={IEEE Transactions on Industrial Informatics},
	volume={16},
	number={4},
	pages={2335--2345},
	year={2019},
	publisher={IEEE}
}

@article{tii3,
	title={Parametric primitive analysis of cad sketches with vision transformer},
	author={Wang, Xiaogang and Wang, Liang and Wu, Hongyu and Xiao, Guoqiang and Xu, Kai},
	journal={IEEE Transactions on Industrial Informatics},
	volume={20},
	number={10},
	pages={12041--12050},
	year={2024},
	publisher={IEEE}
}

\end{document}